\documentclass[letterpaper,12pt,english]{article}
\usepackage[utf8]{inputenc}
\DeclareUnicodeCharacter{00A0}{\nobreakspace}
\usepackage{cmap}
\usepackage[T1]{fontenc}
\usepackage{babel}
\usepackage{times}
\usepackage[Bjarne]{fncychap}
\usepackage{longtable}
\usepackage{sphinx}
\usepackage{multirow}
\usepackage{eqparbox}
\usepackage{amsfonts}

\addto\captionsenglish{}
\addto\captionsenglish{}
\SetupFloatingEnvironment{literal-block}{name=Listing }

  \pdfoutput=1
  \usepackage{cmap}
  \usepackage{amssymb}
  \newcommand{\chapter}[1]{}  
  \newcommand{\ignore}[1]{}  

\title{{COCO}: The Experimental Procedure}
\date{May 19, 2016}
\release{0.9}
\date{\vspace{-1ex}}\author{Nikolaus Hansen$^{1,2}$, 
      Tea Tu\v{s}ar$^3$, 
      Olaf Mersmann$^4$, 
      Anne Auger$^{1,2}$, 
      Dimo Brockhoff$^3$
  \\
    $^1$Inria, research centre Saclay, France
  \\
   $^2$Universit\'e Paris-Saclay, LRI, France
  \\
    $^3$Inria, research centre Lille, France
  \\
    $^4$TU Dortmund University, Chair of Computational Statistics, Germany
    }

\renewcommand{\releasename}{Release}

\makeindex

\makeatletter
\def\PYG@reset{\let\PYG@it=\relax \let\PYG@bf=\relax%
    \let\PYG@ul=\relax \let\PYG@tc=\relax%
    \let\PYG@bc=\relax \let\PYG@ff=\relax}
\def\PYG@tok#1{\csname PYG@tok@#1\endcsname}
\def\PYG@toks#1+{\ifx\relax#1\empty\else%
    \PYG@tok{#1}\expandafter\PYG@toks\fi}
\def\PYG@do#1{\PYG@bc{\PYG@tc{\PYG@ul{%
    \PYG@it{\PYG@bf{\PYG@ff{#1}}}}}}}
\def\PYG#1#2{\PYG@reset\PYG@toks#1+\relax+\PYG@do{#2}}

\expandafter\def\csname PYG@tok@gd\endcsname{\def\PYG@tc##1{\textcolor[rgb]{0.63,0.00,0.00}{##1}}}
\expandafter\def\csname PYG@tok@gu\endcsname{\let\PYG@bf=\textbf\def\PYG@tc##1{\textcolor[rgb]{0.50,0.00,0.50}{##1}}}
\expandafter\def\csname PYG@tok@gt\endcsname{\def\PYG@tc##1{\textcolor[rgb]{0.00,0.27,0.87}{##1}}}
\expandafter\def\csname PYG@tok@gs\endcsname{\let\PYG@bf=\textbf}
\expandafter\def\csname PYG@tok@gr\endcsname{\def\PYG@tc##1{\textcolor[rgb]{1.00,0.00,0.00}{##1}}}
\expandafter\def\csname PYG@tok@cm\endcsname{\let\PYG@it=\textit\def\PYG@tc##1{\textcolor[rgb]{0.25,0.50,0.56}{##1}}}
\expandafter\def\csname PYG@tok@vg\endcsname{\def\PYG@tc##1{\textcolor[rgb]{0.73,0.38,0.84}{##1}}}
\expandafter\def\csname PYG@tok@vi\endcsname{\def\PYG@tc##1{\textcolor[rgb]{0.73,0.38,0.84}{##1}}}
\expandafter\def\csname PYG@tok@mh\endcsname{\def\PYG@tc##1{\textcolor[rgb]{0.13,0.50,0.31}{##1}}}
\expandafter\def\csname PYG@tok@cs\endcsname{\def\PYG@tc##1{\textcolor[rgb]{0.25,0.50,0.56}{##1}}\def\PYG@bc##1{\setlength{\fboxsep}{0pt}\colorbox[rgb]{1.00,0.94,0.94}{\strut ##1}}}
\expandafter\def\csname PYG@tok@ge\endcsname{\let\PYG@it=\textit}
\expandafter\def\csname PYG@tok@vc\endcsname{\def\PYG@tc##1{\textcolor[rgb]{0.73,0.38,0.84}{##1}}}
\expandafter\def\csname PYG@tok@il\endcsname{\def\PYG@tc##1{\textcolor[rgb]{0.13,0.50,0.31}{##1}}}
\expandafter\def\csname PYG@tok@go\endcsname{\def\PYG@tc##1{\textcolor[rgb]{0.20,0.20,0.20}{##1}}}
\expandafter\def\csname PYG@tok@cp\endcsname{\def\PYG@tc##1{\textcolor[rgb]{0.00,0.44,0.13}{##1}}}
\expandafter\def\csname PYG@tok@gi\endcsname{\def\PYG@tc##1{\textcolor[rgb]{0.00,0.63,0.00}{##1}}}
\expandafter\def\csname PYG@tok@gh\endcsname{\let\PYG@bf=\textbf\def\PYG@tc##1{\textcolor[rgb]{0.00,0.00,0.50}{##1}}}
\expandafter\def\csname PYG@tok@ni\endcsname{\let\PYG@bf=\textbf\def\PYG@tc##1{\textcolor[rgb]{0.84,0.33,0.22}{##1}}}
\expandafter\def\csname PYG@tok@nl\endcsname{\let\PYG@bf=\textbf\def\PYG@tc##1{\textcolor[rgb]{0.00,0.13,0.44}{##1}}}
\expandafter\def\csname PYG@tok@nn\endcsname{\let\PYG@bf=\textbf\def\PYG@tc##1{\textcolor[rgb]{0.05,0.52,0.71}{##1}}}
\expandafter\def\csname PYG@tok@no\endcsname{\def\PYG@tc##1{\textcolor[rgb]{0.38,0.68,0.84}{##1}}}
\expandafter\def\csname PYG@tok@na\endcsname{\def\PYG@tc##1{\textcolor[rgb]{0.25,0.44,0.63}{##1}}}
\expandafter\def\csname PYG@tok@nb\endcsname{\def\PYG@tc##1{\textcolor[rgb]{0.00,0.44,0.13}{##1}}}
\expandafter\def\csname PYG@tok@nc\endcsname{\let\PYG@bf=\textbf\def\PYG@tc##1{\textcolor[rgb]{0.05,0.52,0.71}{##1}}}
\expandafter\def\csname PYG@tok@nd\endcsname{\let\PYG@bf=\textbf\def\PYG@tc##1{\textcolor[rgb]{0.33,0.33,0.33}{##1}}}
\expandafter\def\csname PYG@tok@ne\endcsname{\def\PYG@tc##1{\textcolor[rgb]{0.00,0.44,0.13}{##1}}}
\expandafter\def\csname PYG@tok@nf\endcsname{\def\PYG@tc##1{\textcolor[rgb]{0.02,0.16,0.49}{##1}}}
\expandafter\def\csname PYG@tok@si\endcsname{\let\PYG@it=\textit\def\PYG@tc##1{\textcolor[rgb]{0.44,0.63,0.82}{##1}}}
\expandafter\def\csname PYG@tok@s2\endcsname{\def\PYG@tc##1{\textcolor[rgb]{0.25,0.44,0.63}{##1}}}
\expandafter\def\csname PYG@tok@nt\endcsname{\let\PYG@bf=\textbf\def\PYG@tc##1{\textcolor[rgb]{0.02,0.16,0.45}{##1}}}
\expandafter\def\csname PYG@tok@nv\endcsname{\def\PYG@tc##1{\textcolor[rgb]{0.73,0.38,0.84}{##1}}}
\expandafter\def\csname PYG@tok@s1\endcsname{\def\PYG@tc##1{\textcolor[rgb]{0.25,0.44,0.63}{##1}}}
\expandafter\def\csname PYG@tok@ch\endcsname{\let\PYG@it=\textit\def\PYG@tc##1{\textcolor[rgb]{0.25,0.50,0.56}{##1}}}
\expandafter\def\csname PYG@tok@m\endcsname{\def\PYG@tc##1{\textcolor[rgb]{0.13,0.50,0.31}{##1}}}
\expandafter\def\csname PYG@tok@gp\endcsname{\let\PYG@bf=\textbf\def\PYG@tc##1{\textcolor[rgb]{0.78,0.36,0.04}{##1}}}
\expandafter\def\csname PYG@tok@sh\endcsname{\def\PYG@tc##1{\textcolor[rgb]{0.25,0.44,0.63}{##1}}}
\expandafter\def\csname PYG@tok@ow\endcsname{\let\PYG@bf=\textbf\def\PYG@tc##1{\textcolor[rgb]{0.00,0.44,0.13}{##1}}}
\expandafter\def\csname PYG@tok@sx\endcsname{\def\PYG@tc##1{\textcolor[rgb]{0.78,0.36,0.04}{##1}}}
\expandafter\def\csname PYG@tok@bp\endcsname{\def\PYG@tc##1{\textcolor[rgb]{0.00,0.44,0.13}{##1}}}
\expandafter\def\csname PYG@tok@c1\endcsname{\let\PYG@it=\textit\def\PYG@tc##1{\textcolor[rgb]{0.25,0.50,0.56}{##1}}}
\expandafter\def\csname PYG@tok@o\endcsname{\def\PYG@tc##1{\textcolor[rgb]{0.40,0.40,0.40}{##1}}}
\expandafter\def\csname PYG@tok@kc\endcsname{\let\PYG@bf=\textbf\def\PYG@tc##1{\textcolor[rgb]{0.00,0.44,0.13}{##1}}}
\expandafter\def\csname PYG@tok@c\endcsname{\let\PYG@it=\textit\def\PYG@tc##1{\textcolor[rgb]{0.25,0.50,0.56}{##1}}}
\expandafter\def\csname PYG@tok@mf\endcsname{\def\PYG@tc##1{\textcolor[rgb]{0.13,0.50,0.31}{##1}}}
\expandafter\def\csname PYG@tok@err\endcsname{\def\PYG@bc##1{\setlength{\fboxsep}{0pt}\fcolorbox[rgb]{1.00,0.00,0.00}{1,1,1}{\strut ##1}}}
\expandafter\def\csname PYG@tok@mb\endcsname{\def\PYG@tc##1{\textcolor[rgb]{0.13,0.50,0.31}{##1}}}
\expandafter\def\csname PYG@tok@ss\endcsname{\def\PYG@tc##1{\textcolor[rgb]{0.32,0.47,0.09}{##1}}}
\expandafter\def\csname PYG@tok@sr\endcsname{\def\PYG@tc##1{\textcolor[rgb]{0.14,0.33,0.53}{##1}}}
\expandafter\def\csname PYG@tok@mo\endcsname{\def\PYG@tc##1{\textcolor[rgb]{0.13,0.50,0.31}{##1}}}
\expandafter\def\csname PYG@tok@kd\endcsname{\let\PYG@bf=\textbf\def\PYG@tc##1{\textcolor[rgb]{0.00,0.44,0.13}{##1}}}
\expandafter\def\csname PYG@tok@mi\endcsname{\def\PYG@tc##1{\textcolor[rgb]{0.13,0.50,0.31}{##1}}}
\expandafter\def\csname PYG@tok@kn\endcsname{\let\PYG@bf=\textbf\def\PYG@tc##1{\textcolor[rgb]{0.00,0.44,0.13}{##1}}}
\expandafter\def\csname PYG@tok@cpf\endcsname{\let\PYG@it=\textit\def\PYG@tc##1{\textcolor[rgb]{0.25,0.50,0.56}{##1}}}
\expandafter\def\csname PYG@tok@kr\endcsname{\let\PYG@bf=\textbf\def\PYG@tc##1{\textcolor[rgb]{0.00,0.44,0.13}{##1}}}
\expandafter\def\csname PYG@tok@s\endcsname{\def\PYG@tc##1{\textcolor[rgb]{0.25,0.44,0.63}{##1}}}
\expandafter\def\csname PYG@tok@kp\endcsname{\def\PYG@tc##1{\textcolor[rgb]{0.00,0.44,0.13}{##1}}}
\expandafter\def\csname PYG@tok@w\endcsname{\def\PYG@tc##1{\textcolor[rgb]{0.73,0.73,0.73}{##1}}}
\expandafter\def\csname PYG@tok@kt\endcsname{\def\PYG@tc##1{\textcolor[rgb]{0.56,0.13,0.00}{##1}}}
\expandafter\def\csname PYG@tok@sc\endcsname{\def\PYG@tc##1{\textcolor[rgb]{0.25,0.44,0.63}{##1}}}
\expandafter\def\csname PYG@tok@sb\endcsname{\def\PYG@tc##1{\textcolor[rgb]{0.25,0.44,0.63}{##1}}}
\expandafter\def\csname PYG@tok@k\endcsname{\let\PYG@bf=\textbf\def\PYG@tc##1{\textcolor[rgb]{0.00,0.44,0.13}{##1}}}
\expandafter\def\csname PYG@tok@se\endcsname{\let\PYG@bf=\textbf\def\PYG@tc##1{\textcolor[rgb]{0.25,0.44,0.63}{##1}}}
\expandafter\def\csname PYG@tok@sd\endcsname{\let\PYG@it=\textit\def\PYG@tc##1{\textcolor[rgb]{0.25,0.44,0.63}{##1}}}

\makeatother

\begin{document}

\maketitle
\phantomsection\label{index::doc}

\chapter{CHAPTERTITLE}
\label{index:coco-the-experimental-procedure}\label{index:chaptertitle}
\begin{abstract}
We present a budget-free experimental setup and procedure for benchmarking numerical
optimization algorithms in a black-box scenario.
This procedure can be applied with the \href{https://github.com/numbbo/coco}{COCO} benchmarking platform.
We describe initialization of and input to the algorithm and touch upon the
relevance of termination and restarts.
\end{abstract}\tableofcontents 
\newpage

\section{Introduction}
\label{index:introduction}
Based on \phantomsection\label{index:id4}{\hyperref[index:han2009]{\emph{{[}HAN2009{]}}}} and \phantomsection\label{index:id5}{\hyperref[index:han2010]{\emph{{[}HAN2010{]}}}}, we describe a comparatively simple experimental
setup for \emph{black-box optimization benchmarking}. We recommend to use this procedure
within the \href{https://github.com/numbbo/coco}{COCO} platform \phantomsection\label{index:id6}{\hyperref[index:han2016co]{\emph{{[}HAN2016co{]}}}}.\footnote[1]{
The \href{https://github.com/numbbo/coco}{COCO} platform provides
several (single and bi-objective) \emph{test suites} with a collection of
black-box optimization problems of different dimensions to be
minimized. \href{https://github.com/numbbo/coco}{COCO} automatically collects the relevant data to display
the performance results after a post-processing is applied.
}

Our \textbf{central measure of performance}, to which the experimental procedure is
adapted, is the number of calls to the objective function to reach a
certain solution quality (function value or \(f\)-value or indicator
value), also denoted as runtime.

\subsection{Terminology}
\label{index:terminology}\begin{description}
\item[{\emph{function}}] \leavevmode
We talk about an objective \emph{function} \(f\) as a parametrized mapping
\(\mathbb{R}^n\to\mathbb{R}^m\) with scalable input space, that is,
\(n\) is not (yet) determined, and usually \(m\in\{1,2\}\).
Functions are parametrized such that different \emph{instances} of the
``same'' function are available, e.g. translated or shifted versions.

\item[{\emph{problem}}] \leavevmode
We talk about a \emph{problem}, \href{http://numbbo.github.io/coco-doc/C/coco\_8h.html\#a408ba01b98c78bf5be3df36562d99478}{\code{coco\_problem\_t}}, as a specific \emph{function
instance} on which the optimization algorithm is run. Specifically, a problem
can be described as the triple \code{(dimension, function, instance)}. A problem
can be evaluated and returns an \(f\)-value or -vector.
In the context of performance
assessment, a target \(f\)- or indicator-value
is attached to each problem. That is, a target value is added to the
above triple to define a single problem in this case.

\item[{\emph{runtime}}] \leavevmode
We define \emph{runtime}, or \emph{run-length} \phantomsection\label{index:id8}{\hyperref[index:hoo1998]{\emph{{[}HOO1998{]}}}}
as the \emph{number of evaluations}
conducted on a given problem, also referred to as number of \emph{function} evaluations.
Our central performance measure is the runtime until a given target value
is hit \phantomsection\label{index:id9}{\hyperref[index:han2016perf]{\emph{{[}HAN2016perf{]}}}}.

\item[{\emph{suite}}] \leavevmode
A test- or benchmark-suite is a collection of problems, typically between
twenty and a hundred, where the number of objectives \(m\) is fixed.

\end{description}

\section{Conducting the Experiment}
\label{index:conducting-the-experiment}
The optimization algorithm to be benchmarked is run on each problem of the
given test suite once. On each problem, the very same algorithm with the same
parameter setting, the same initialzation procedure, the same budget, the same
termination and/or restart criteria etc. is used.
There is no prescribed minimal or maximal allowed budget, the benchmarking
setup is \emph{budget-free}.
The longer the experiment, the more data are available to assess the performance
accurately.
See also Section {\hyperref[index:sec\string-budget]{\emph{Budget, Termination Criteria, and Restarts}}}.

\subsection{Initialization and Input to the Algorithm}
\label{index:initialization-and-input-to-the-algorithm}\label{index:sec-input}
An algorithm can use the following input information from each problem.
At any time:
\begin{description}
\item[{\emph{Input and output dimensions}}] \leavevmode
as a defining interface to the problem, specifically:
\begin{itemize}
\item {} 
The search space (input) dimension via \href{http://numbbo.github.io/coco-doc/C/coco\_8h.html\#a0dabf3e4f5630d08077530a1341f13ab}{\code{coco\_problem\_get\_dimension}},

\item {} 
The number of objectives via \href{http://numbbo.github.io/coco-doc/C/coco\_8h.html\#ab0d1fcc7f592c283f1e67cde2afeb60a}{\code{coco\_problem\_get\_number\_of\_objectives}},
which is the ``output'' dimension of \href{http://numbbo.github.io/coco-doc/C/coco\_8h.html\#aabbc02b57084ab069c37e1c27426b95c}{\code{coco\_evaluate\_function}}.
All functions of a single benchmark suite have the same number
of objectives, currently either one or two.

\item {} 
The number of constraints via \href{http://numbbo.github.io/coco-doc/C/coco\_8h.html\#ad5c7b0889170a105671a14c8383fbb22}{\code{coco\_problem\_get\_number\_of\_constraints}},
which is the ``output'' dimension of \href{http://numbbo.github.io/coco-doc/C/coco\_8h.html\#ab5cce904e394349ec1be1bcdc35967fa}{\code{coco\_evaluate\_constraint}}. \emph{All}
problems of a single benchmark suite have either no constraints, or
one or more constraints.

\end{itemize}

\item[{\emph{Search domain of interest}}] \leavevmode
defined from \href{http://numbbo.github.io/coco-doc/C/coco\_8h.html\#a29c89e039494ae8b4f8e520cba1eb154}{\code{coco\_problem\_get\_largest\_values\_of\_interest}} and \href{http://numbbo.github.io/coco-doc/C/coco\_8h.html\#a4ea6c067adfa866b0179329fe9b7c458}{\code{coco\_problem\_get\_smallest\_values\_of\_interest}}. The optimum (or each extremal solution of the Pareto set) lies within the search domain of interest. If the optimizer operates on a bounded domain only, the domain of interest can be interpreted as lower and upper bounds.

\item[{\emph{Feasible (initial) solution}}] \leavevmode
provided by \href{http://numbbo.github.io/coco-doc/C/coco\_8h.html\#ac5a44845acfadd7c5cccb9900a566b32}{\code{coco\_problem\_get\_initial\_solution}}.

\end{description}

The initial state of the optimization algorithm and its parameters shall only be based on
these input values. The initial algorithm setting is considered as part of
the algorithm and must therefore follow the same procedure for all problems of the
suite. The problem identifier or the positioning of the problem in the suite or
any (other) known characteristics of the problem are not
allowed as input to the algorithm, see also Section
{\hyperref[index:sec\string-tuning]{\emph{Parameter Setting and Tuning of Algorithms}}}.

During an optimization run, the following (new) information is available to
the algorithm:
\begin{enumerate}
\item {} 
The result, i.e., the \(f\)-value(s) from evaluating the problem
at a given search point
via \href{http://numbbo.github.io/coco-doc/C/coco\_8h.html\#aabbc02b57084ab069c37e1c27426b95c}{\code{coco\_evaluate\_function}}.

\item {} 
The result from evaluating the constraints of the problem at a
given search point via \href{http://numbbo.github.io/coco-doc/C/coco\_8h.html\#ab5cce904e394349ec1be1bcdc35967fa}{\code{coco\_evaluate\_constraint}}.

\item {} 
The result of \href{http://numbbo.github.io/coco-doc/C/coco\_8h.html\#a1164d85fd641ca48046b943344ae9069}{\code{coco\_problem\_final\_target\_hit}}, which can be used
to terminate a run conclusively without changing the performance assessment
in any way. Currently, if the number of objectives \(m > 1\), this
function returns always zero.

\end{enumerate}

The number of evaluations of the problem and/or constraints are the search
costs, also referred to as \emph{runtime}, and used for the performance
assessment of the algorithm.\footnote[2]{
\href{http://numbbo.github.io/coco-doc/C/coco\_8h.html\#a6ad88cdba2ffd15847346d594974067f}{\code{coco\_problem\_get\_evaluations(const coco\_problem\_t * problem)}} is a
convenience function that returns the number of evaluations done on \code{problem}.
Because this information is available to the optimization algorithm anyway,
the convenience function might be used additionally.
}

\subsection{Budget, Termination Criteria, and Restarts}
\label{index:sec-budget}\label{index:budget-termination-criteria-and-restarts}
Algorithms and/or setups with any budget of function evaluations are
eligible, the benchmarking setup is \emph{budget-free}.
We consider termination criteria to be part of the benchmarked algorithm.
The choice of termination is a relevant part of the algorithm.
On the one hand, allowing a larger number of function evaluations increases the chance to find solutions with better quality. On the other hand, a timely
termination of stagnating runs can improve the performance, as these evaluations
can be used more effectively.

To exploit a large(r) number of function evaluations effectively, we encourage to
use \textbf{independent restarts}\footnote[3]{
The \href{https://github.com/numbbo/coco}{COCO} platform provides example code implementing independent restarts.
}, in particular for algorithms which terminate
naturally within a comparatively small budget.
Independent restarts are a natural way to approach difficult optimization
problems and do not change the central performance measure used in \href{https://github.com/numbbo/coco}{COCO} (hence it is budget-free),
however,
independent restarts improve the reliability, comparability\footnote[4]{
Algorithms are only comparable up to the smallest budget given to
any of them.
}, precision, and ``visibility'' of the measured results.

Moreover, any \textbf{multistart procedure} (which relies on an interim termination of the algorithm) is encouraged.
Multistarts may not be independent as they can feature a parameter sweep (e.g., increasing population size \phantomsection\label{index:id15}{\hyperref[index:har1999]{\emph{{[}HAR1999{]}}}} \phantomsection\label{index:id16}{\hyperref[index:aug2005]{\emph{{[}AUG2005{]}}}}), can be based on the outcome of the previous starts, and/or feature a systematic change of the initial conditions for the algorithm.

After a multistart procedure has been established, a recommended procedure is
to use a budget proportional to the dimension, \(k\times n\), and run
repeated experiments with increase \(k\), e.g. like
\(3, 10, 30, 100, 300,\dots\), which is a good compromise between
availability of the latest results and computational overhead.

An algorithm can be conclusively terminated if
\href{http://numbbo.github.io/coco-doc/C/coco\_8h.html\#a1164d85fd641ca48046b943344ae9069}{\code{coco\_problem\_final\_target\_hit}} returns 1.\footnote[5]{
For the \code{bbob-biobj} suite this is however currently never the case.
} This saves CPU cycles without
affecting the performance assessment, because there is no target left to hit.

\section{Parameter Setting and Tuning of Algorithms}
\label{index:sec-tuning}\label{index:parameter-setting-and-tuning-of-algorithms}
Any tuning of algorithm parameters to the test suite should be described and
\emph{the approximate overall number of tested parameter settings or algorithm
variants and the approximate overall invested budget should be given}.

The only recommended tuning procedure is the verification that \textbf{termination
conditions} of the algorithm are suited to the given testbed and, in case,
tuning of termination parameters.\footnote[6]{
For example in the single objective case, care should be
taken to apply termination conditions that allow to hit the final target on
the most basic functions, like the sphere function \(f_1\), that is on the
problems 0, 360, 720, 1080, 1440, and 1800 of the \code{bbob} suite.

In our experience, numerical optimization software frequently terminates
too early by default, while evolutionary computation software often
terminates too late by default.
}
Too early or too late termination can be identified and adjusted comparatively
easy.
This is also a useful prerequisite for allowing restarts to become more effective.

On all functions the very same parameter setting must be used (which might
well depend on the dimensionality, see Section {\hyperref[index:sec\string-input]{\emph{Initialization and Input to the Algorithm}}}). That means,
the \emph{a priori} use of function-dependent parameter settings is prohibited
(since 2012).  The function ID or any function characteristics (like
separability, multi-modality, ...) cannot be considered as input parameter to
the algorithm.

On the other hand, benchmarking different parameter settings as ``different
algorithms'' on the entire test suite is encouraged.

\section{Time Complexity Experiment}
\label{index:time-complexity-experiment}
In order to get a rough measurement of the time complexity of the algorithm, the
wall-clock or CPU time should be measured when running the algorithm on the
benchmark suite. The chosen setup should reflect a ``realistic average
scenario''.\footnote[7]{
The example experiment code provides the timing output measured over all
problems of a single dimension by default. It also can be used to make a record
of the same timing experiment with ``pure random search'', which can serve as
additional base-line data. On the \code{bbob} test suite, also only the
first instance of the Rosenbrock function \(f_8\) had been used for this
experiment previously, that is, the suite indices 105, 465, 825, 1185, 1545,
1905.
}
The \emph{time divided by the number of function evaluations shall be presented
separately for each dimension}. The chosen setup, coding language, compiler and
computational architecture for conducting these experiments should be given.
\section*{Acknowledgments}
The authors would like to thank Raymond Ros, Steffen Finck, Marc Schoenauer,
Petr Posik and Dejan Tušar for their many invaluable contributions to this work.

This work was support by the grant ANR-12-MONU-0009 (NumBBO)
of the French National Research Agency.

\renewcommand{\indexname}{Index}
\printindex
\end{document}